\documentclass[10pt,twocolumn,letterpaper]{article}

\usepackage[pagenumbers]{iccv}   %

\PassOptionsToPackage{dvipsnames,table}{xcolor}

\usepackage{wrapfig}
\usepackage{colortbl}
\usepackage[cjk]{kotex}

\usepackage{lipsum}
\usepackage[normalem]{ulem}

\usepackage{graphicx}
\usepackage{multirow}
\usepackage{booktabs}
\usepackage{array}
\usepackage[ruled,linesnumbered]{algorithm2e}
\usepackage{amsmath,amssymb}
\usepackage{enumitem}

\definecolor{iccvblue}{rgb}{0.21,0.49,0.74}
\usepackage[pagebackref,breaklinks,colorlinks,allcolors=iccvblue]{hyperref}

\usepackage{xcolor}
\usepackage{listings}
\usepackage{algpseudocode}
\lstdefinestyle{pythonstyle}{
    language=Python,
    basicstyle=\ttfamily\footnotesize,
    keywordstyle=\color{keywordblue}\bfseries,
    commentstyle=\color{commentgreen},
    stringstyle=\color{stringred},
    breaklines=true,
    showstringspaces=false,
    tabsize=2
}
\definecolor{commentgreen}{rgb}{0,0.5,0}  %
\definecolor{keywordblue}{rgb}{0,0,0.8}   %
\definecolor{stringred}{rgb}{0.6,0,0}     %

\def\paperID{52} %
\def\confName{ICCV}
\def\confYear{2025}

\title{Revisiting Reliability in the Reasoning-based Pose Estimation Benchmark}

\author{%
  Junsu Kim$^{1,2}$%
  \qquad
  Naeun Kim$^{1}$\thanks{Equal contribution.}\,\qquad %
  Jaeho Lee$^{1}$\footnotemark[1]\,\qquad %
  Incheol Park$^{1}$\footnotemark[1] \qquad \\
  Dongyoon Han$^{3}$\thanks{Corresponding author.}\qquad%
  Seungryul Baek$^{1}$\footnotemark[2]%
  \\[4pt]
  $^{1}$UNIST \qquad
  $^{2}$NVIDIA Foundation Models Lab, MODULABS \qquad
  $^{3}$NAVER AI Lab
}

\begin{document}

\maketitle
\begin{abstract}
    The reasoning-based pose estimation (RPE) benchmark has emerged as a widely adopted evaluation standard for pose-aware multimodal large language models (MLLMs). Despite its significance, we identified critical reproducibility and benchmark-quality issues that hinder fair and consistent quantitative evaluations. Most notably, the benchmark utilizes different image indices from those of the original 3DPW dataset, forcing researchers into tedious and error-prone manual matching processes to obtain accurate ground-truth (GT) annotations for quantitative metrics (\eg, MPJPE, PA-MPJPE). Furthermore, our analysis reveals several inherent benchmark-quality limitations, including significant image redundancy, scenario imbalance, overly simplistic poses, and ambiguous textual descriptions, collectively undermining reliable evaluations across diverse scenarios. To alleviate manual effort and enhance reproducibility, we carefully refined the GT annotations through meticulous visual matching and publicly release these refined annotations as an open-source resource, thereby promoting consistent quantitative evaluations and facilitating future advancements in human pose-aware multimodal reasoning.
\end{abstract}
    
\section{Introduction}
\label{sec:introduction}

Human-centric AI increasingly demands semantic understanding beyond mere geometric accuracy in human pose estimation, especially for downstream tasks such as augmented reality (AR) coaching, assistive robotics, and social-scene interpretation. Traditional approaches—e.g., HMR~\cite{kanazawa2018HMR1.0}, ViTPose~\cite{xu2022vitpose}, Pose2Mesh~\cite{choi2020pose2mesh}, CLIFF~\cite{li2022cliff}—excel at localizing joints and fitting parametric models like SMPL~\cite{SMPL}, yet inherently lack linguistic context, limiting their capacity to interpret detailed human intentions and complex interactions.
To this end, early attempts to narrow this semantic gap, such as PoseScript~\cite{PoseScript} and PoseFix~\cite{delmas2023posefix}, attached text descriptions to pose data. 
While pioneering works such as PoseScript and PoseFix demonstrated the value of coupling language cues with pose data, they remained largely joint-centric, relying on rigid, hand-crafted templates fine-tuned for pose description. Because these templates lack the breadth and adaptability of full linguistic grounding, they can not capture the nuances of natural human language, severely constraining their ability to reason about diverse, real-world scenarios.
\begin{figure}[t]
    \centering  
    \vspace{-1.5em}
    \includegraphics[width=\linewidth]{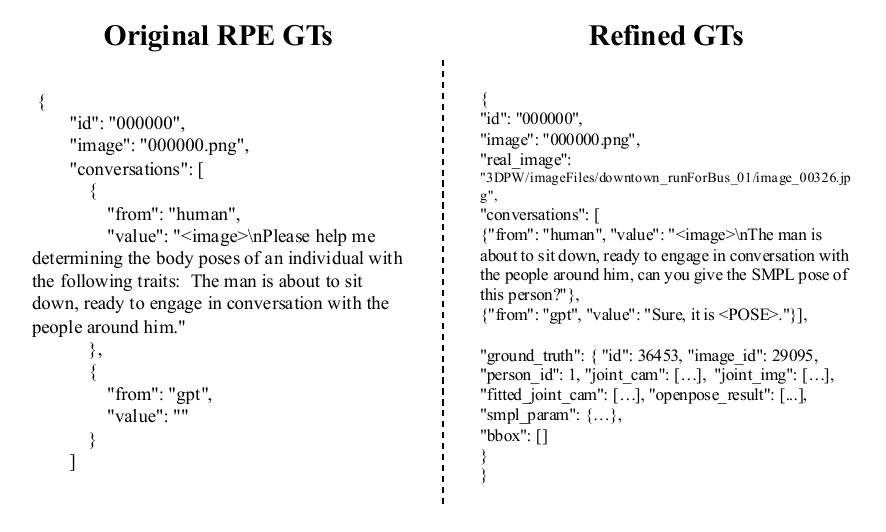}
    \caption{\textbf{Original RPE ground truths (\emph{left}) vs.\ Refined ground truths (\emph{right}).} Our refined version clearly links each example to the correct original 3DPW frame path (\texttt{real\_image}) and includes essential annotations for precise quantitative evaluations, such as SMPL~\cite{SMPL} parameters and 3D joint coordinates (\texttt{joint\_cam}) for MPJPE and PA-MPJPE. Additional metadata is also provided to support accurate and convenient mapping. We publicly release the JSON file of these refined ground truths to facilitate reproducible evaluations.}
    \label{fig:rpe_gt_cmp}
\end{figure}
Concurrently, broader efforts within the vision–language community have emerged, supplying large language models (LLMs) with diverse visual and non-visual inputs—such as images, videos, depth maps, and text—and demonstrating impressive multimodal reasoning capabilities in various tasks~\cite{li2023blip2,liu2024llavanext,li2023videochat,zhang2023videollama}. Riding this wave, recent human pose-aware multimodal LLMs (MLLMs)~\cite{PoseGPT,li2025unipose,feng2025posellava,lin2024chathuman} have shown significant promise in bridging the semantic gap in human pose understanding by seamlessly integrating pose estimation and visual perception with comprehensive linguistic common sense.

\begin{figure*}[t]
\centering
\includegraphics[width=\textwidth]{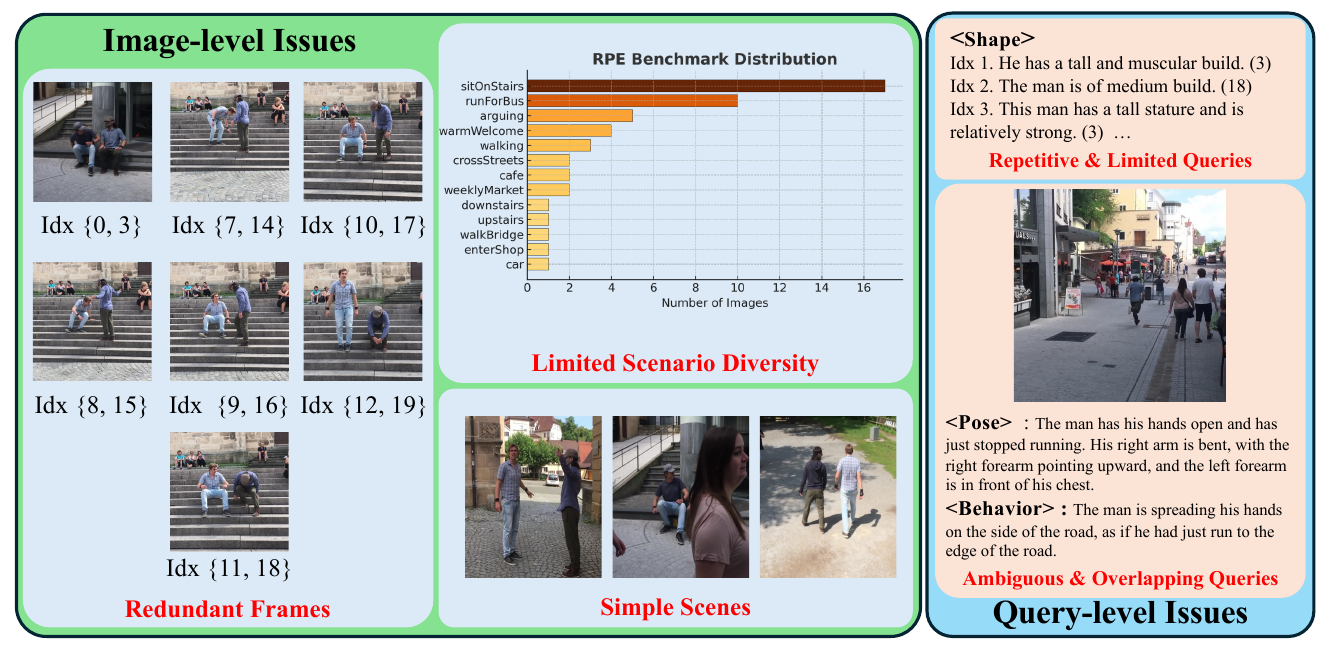}
\caption{\textbf{Issues of the existing RPE benchmark.} We summarize potential internal issues within the RPE benchmark that could negatively impact its effectiveness for robust evaluation. \textit{Image-level} issues include redundant frames (\eg, indices 
\{0, 3\} indicate nearly identical or duplicate images), overly simplistic scenes, and limited scenario diversity. \textit{Query-level} issues are identified in textual descriptions across three attributes: \textbf{\textless Shape\textgreater} (physical characteristics of individuals), \textbf{\textless Pose\textgreater} (specific body poses and joint configurations), and \textbf{\textless Behavior\textgreater} (contextual actions or interactions within scenes). These query-level issues include repetitive textual prompts (\eg, ``(3)", ``(18)" indicate the number of same prompts in \textbf{\textless Shape\textgreater}), as well as significant ambiguity and overlapping descriptions between the \textbf{\textless Pose\textgreater} and \textbf{\textless Behavior\textgreater} attributes.}
\label{fig:rpe_quality_issues}
\end{figure*}
To rigorously evaluate these emerging human pose-aware MLLMs, the reasoning-based pose estimation (RPE) benchmark, introduced by ChatPose~\cite{ChatPose}, represents the first public attempt designed specifically to assess models' reasoning capabilities in identifying a particular person from linguistic and visual descriptions, and generating accurate SMPL pose parameters for that individual. Rapidly becoming the de facto standard, RPE has significantly influenced related research directions, being concurrently adopted by works such as UniPose~\cite{li2025unipose} and ChatHuman~\cite{lin2024chathuman}.

However, despite RPE's widespread adoption and considerable influence, potential issues related to reproducibility and benchmark quality might undermine the reliability of its evaluations, possibly leading to inaccurate conclusions. Specifically, since the RPE benchmark does not provide accurate ground-truth annotations aligned with the original 3DPW~\cite{von20183dpw} dataset, researchers aiming to perform quantitative evaluations must manually conduct one-to-one visual matching between RPE images and corresponding original 3DPW frames to retrieve the necessary annotations (\cref{fig:rpe_gt_cmp}, \textit{left}). Such manual matching is labor-intensive, error-prone, and significantly undermines the reproducibility and reliability of quantitative analyses. Additionally, intrinsic benchmark limitations—including redundant imagery, scenario imbalances, overly simplistic scenes, and ambiguous textual queries—could further compromise benchmark reliability, potentially causing subsequent studies to favor qualitative assessments over essential quantitative analyses (\cref{fig:rpe_quality_issues}). Addressing these challenges is critical for enabling rigorous, consistent evaluations of human pose-aware multimodal models and fostering meaningful progress within the research community.

In this paper, we systematically investigate and address critical technical and benchmark-quality shortcomings within the RPE benchmark. To overcome reproducibility challenges associated with labor-intensive manual matching processes, we carefully conducted a detailed visual alignment of each RPE instance with its original 3DPW frame, subsequently releasing these precisely refined ground-truth annotations as an open-source resource (\cref{fig:rpe_gt_cmp}, \textit{right}). Furthermore, we provide comprehensive documentation of intrinsic benchmark issues to guide future improvements, thereby facilitating transparent, reliable, and rigorous quantitative evaluations that ultimately advance the development of pose-aware multimodal large language models.

\section{Related Work}
\label{sec:related_work}
\subsection{Multimodal large language models}
\label{subsec:rw_mllms}
Multimodal large language models (MLLMs) significantly extended the powerful reasoning capabilities of language models into visual domains. Numerous open-source MLLMs emerged, typically freezing both the vision encoder and language backbone while learning a lightweight projection (MLP) to align multimodal features. For instance, BLIP-2~\cite{li2023blip2} couples a frozen CLIP-based visual encoder~\cite{radford2021CLIP, Ilharco2021OpenCLIP} to language models (OPT~\cite{zhang2022opt}/Flan-T5~\cite{chung2024FlanT5}) using a learned Q-Former, whereas MiniGPT-4~\cite{zhu2023minigpt} simplifies this with a single linear mapping. Further advancing visual instruction tuning, LLaVA-1.5~\cite{liu2024llava1-5} employs Vicuna~\cite{zheng2023vicuna}, itself a fine-tuned variant of LLaMA~\cite{touvron2023llama}, to enable richer visual dialogues. 
Subsequent efforts broadened beyond static images to diverse input modalities: models like VideoChat~\cite{li2023videochat} and Video-LLaMA~\cite{zhang2023videollama} incorporate temporal visual reasoning from video data, and SpeechGPT~\cite{zhang2023speechgpt} enables conversational understanding from audio input.

\subsection{Human pose-aware MLLMs}
Beyond general-purpose MLLMs, recent studies increasingly focus on integrating explicit human pose information, enabling richer human-centric reasoning. ChatPose~\cite{ChatPose} introduced ``pose tokens," encoding SMPL~\cite{SMPL} parameters within linguistic frameworks, and presented tasks such as speculative pose generation and reasoning-based pose estimation (RPE). Similarly, PoseLLaVA~\cite{feng2025posellava} embeds SMPL-based pose tokens into the LLaVA~\cite{liu2024llavanext} architecture, leveraging token-level cross-attention for language-driven 3D pose manipulation. UniPose~\cite{li2025unipose} proposed a unified ``pose vocabulary" alongside multiple vision encoders (e.g., CLIP~\cite{radford2021CLIP} and PoseViT~\cite{goel2023humans2.0}), supporting zero-shot generalization across diverse pose-related tasks. ChatHuman~\cite{lin2024chathuman} adopts retrieval-augmented strategies, dynamically incorporating external pose and interaction models as expert tools to enhance context-sensitive reasoning. In light of these advances, the RPE benchmark has emerged as a critical tool for evaluating the text-driven reasoning capabilities of human pose-aware MLLMs. However, despite its significance, we identified certain technical and potential quality issues within the benchmark that may compromise reliable evaluations. In this paper, we explicitly highlight these limitations and propose refinements, laying a robust foundation for rigorous and transparent analysis of human pose understanding integrated with LLM-based multimodal reasoning.

\section{Reasoning-based pose estimation}
\label{section:preliminary}
The reasoning-based pose estimation (RPE) benchmark, introduced by ChatPose~\cite{ChatPose}, evaluates MLLMs specifically on their ability to reason about human poses through linguistic and visual contexts. Unlike traditional benchmarks that emphasize geometric accuracy alone, RPE integrates semantic reasoning into the pose estimation task. The RPE dataset comprises 50 multiple-person images selected from the widely used 3DPW dataset~\cite{von20183dpw}. Descriptions for each individual in these images were initially generated using GPT-4V, covering diverse attributes such as \textit{behavior}, \textit{outfits}, \textit{pose}, \textit{shape}, and \textit{overall context}. These automatically generated descriptions underwent manual refinement, resulting in 250 question-answer pairs formatted as:\texttt{<IMAGE> \{person description\}, can you give the SMPL pose of this person?} This explicit query template tests whether MLLMs can effectively leverage linguistic prompts combined with visual cues to estimate accurate human poses, without relying on additional steps such as bounding box extraction. Consequently, the RPE benchmark uniquely measures the ability of human pose-aware MLLMs to utilize comprehensive world knowledge inherent in LLMs, thereby providing crucial insights into their semantic reasoning abilities in pose-related tasks.

\section{Identified limitations}
\label{section:identified_limitations}

Although the RPE benchmark has become influential in evaluating pose-aware MLLMs, we identify several critical issues that significantly compromise its reliability and robustness. These include technical reproducibility challenges (\cref{subsec:technical_issue}, \cref{fig:rpe_gt_cmp}), benchmark-quality limitations (\cref{subsec:quality_issue}, \cref{fig:rpe_quality_issues}), and inherent annotation and preprocessing constraints (\cref{subsec:Inherent-issue}, \cref{fig:issue-others}). Each category is elaborated in the subsequent sections.

\subsection{Reproducibility issues}
\label{subsec:technical_issue}
The primary technical issue in the RPE benchmark is its lack of reproducibility. Specifically, the benchmark assigns custom image indices (\cref{fig:rpe_gt_cmp}, \textit{left}) that differ from the original indices provided in the 3DPW dataset~\cite{von20183dpw}. This discrepancy forces researchers aiming to perform accurate and fair quantitative evaluations to manually perform labor-intensive, error-prone visual matching between each RPE example and its corresponding original 3DPW frame to obtain correct ground-truth (GT) annotations. Despite the seemingly manageable size of the RPE dataset (50 images), manual matching is significantly complicated by the inherent nature of the 3DPW dataset, as it includes frames sampled from extended video sequences containing numerous visually similar images. Moreover, the presence of lengthy frame sequences across each of the 24 testing scenarios in 3DPW further increases the complexity and difficulty of accurately identifying matches. Consequently, this technical limitation has considerably impeded rigorous quantitative analyses in recent studies~\cite{li2025unipose,lin2024chathuman}, posing substantial obstacles to fair comparisons and constraining further advancements in the evaluation of pose-aware MLLMs.

\subsection{Quality issues}
\label{subsec:quality_issue}
In addition to reproducibility challenges, the RPE benchmark suffers from intrinsic quality issues that undermine its robustness as an evaluation standard, as illustrated in \cref{fig:rpe_quality_issues}. We identify four key limitations:

\noindent\textbf{Limited dataset size and redundancy.}
The benchmark consists of only 50 images, inherently limiting its representational diversity and robustness. This issue is compounded by significant redundancy, exemplified by seven pairs of nearly identical or duplicate frames (\eg, indices \{0, 3\}, \{7, 14\}, \etc). Such redundancy substantially reduces the benchmark’s effectiveness in evaluating pose-aware MLLMs' generalization capabilities.

\noindent\textbf{Scenario imbalance.}
Although the original 3DPW dataset encompasses 24 diverse scenarios, the RPE benchmark disproportionately emphasizes a limited subset of scenarios (\eg, ``sitOnStairs," ``runForBus," and ``arguing"), as shown in the scenario distribution chart in \cref{fig:rpe_quality_issues}. This imbalance creates repetitive contexts, similar actions, and recurring individuals, making it difficult to reliably assess a model’s generalization to diverse real-world situations.

\noindent\textbf{Simplistic scenes.}
The benchmark frequently contains trivial scenarios where subjects are simply ``standing" or ``walking," as depicted by the simple scenes examples in \cref{fig:rpe_quality_issues}. Such straightforward cases are typically easily handled by existing vision-language models~\cite{liu2024llava1-5,liu2024llavanext}, underscoring the need for more complex and challenging scenarios that specifically test advanced pose-aware reasoning capabilities.

\begin{figure}[t!]
\centering
\includegraphics[width=\linewidth]{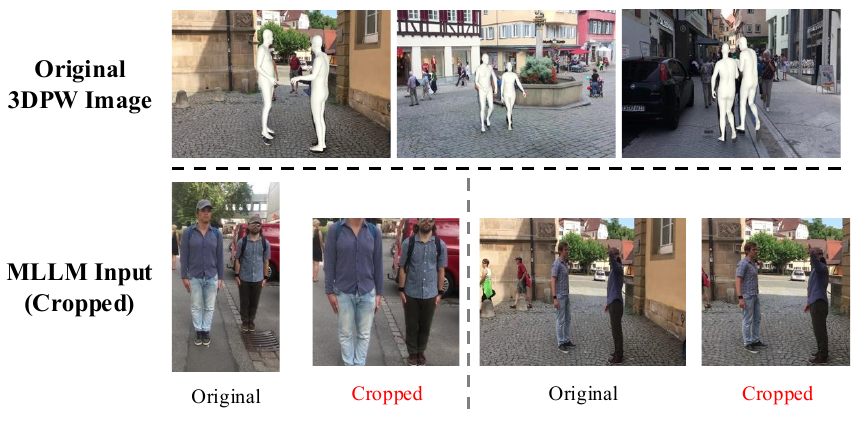}
\caption{\textbf{Additional inherent issues in RPE benchmark.} 
(Top row) Original 3DPW images with SMPL annotations obtained via NeuralAnnot~\cite{Moon_2022_CVPRW_NeuralAnnot}. (Bottom row) Corresponding images after center cropping for MLLM input; the highlighted \textcolor{red}{cropped} illustrates significant cropping of body parts or unintended omission of person.}
\label{fig:issue-others}
\end{figure}
\noindent\textbf{Ambiguous and repetitive queries.}
Textual prompts within the benchmark exhibit significant repetition, particularly within the \textbf{\textless Shape\textgreater} category, limiting meaningful assessments of a model's linguistic understanding and reasoning capabilities across varied contexts. Additionally, descriptions provided under \textbf{\textless Pose\textgreater} and \textbf{\textless Behavior\textgreater} categories often overlap and contain considerable ambiguity. Such unclear prompts significantly increase the likelihood of misinterpretation, especially in multi-person scenes, complicating precise and meaningful evaluations.

Highlighting these quality limitations underscores the critical need to refine and enhance the RPE benchmark for reliable and rigorous evaluation of pose-aware MLLMs.

\subsection{Inherent issues}
\label{subsec:Inherent-issue}
Beyond the limitations discussed above, we identify additional inherent issues, illustrated in~\cref{fig:issue-others}, which affect evaluation robustness.

\noindent\textbf{Incomplete annotations for multi-person scenarios.}
When projecting SMPL ground-truth annotations (obtained via NeuralAnnot~\cite{Moon_2022_CVPRW_NeuralAnnot}) onto the original 3DPW images (\cref{fig:issue-others}, top row), annotations cover at most two individuals per frame (or sometimes cover one person), despite many scenes containing multiple people. This incomplete annotation inherently restricts the representational diversity and poses challenges in comprehensively evaluating model performance within complex, multi-person contexts.

\noindent\textbf{Information loss due to cropping.}
Vision foundation models, widely used by MLLMs, such as CLIP~\cite{radford2021CLIP} or DINOv2~\cite{oquab2023DINOv2}, require fixed-size square image inputs. Consequently, common preprocessing steps, such as center cropping, inadvertently remove critical visual context or partially omit important body regions (\cref{fig:issue-others}, bottom row). This practice can result in evaluation scenarios where the annotated individuals are not fully visible to the models, thereby unintentionally simplifying tasks and potentially introducing performance gains in the RPE benchmark.

\begin{table*}[t]
    \centering
    \resizebox{\linewidth}{!}{%
    \begin{tabular}{l|cccc|c}
    \toprule
        \textbf{Method} & \textbf{Behavior} & \textbf{Shape} & \textbf{Outfit} & \textbf{Pose} & \textbf{Averaged} \\
    \midrule
        ChatPose (\textit{Original})~\cite{ChatPose}  & 307.9 / 102.9 & 269.9 / 103.7 & 265.6 / 102.6 & 277.9 / \textbf{96.0} & 280.3 / 101.3 \\
    \midrule
        ChatPose (fp16)~\cite{ChatPose}  & 243.1 / 105.3 & \textbf{232.8} / 108.4 & 234.6 / 104.1 & \textbf{214.1} / 101.2 & 231.2 / 104.8 \\
        ChatPose (bf16)~\cite{ChatPose}  & \textbf{234.1} / 104.2 & 233.5 / 106.4 & \textbf{234.3} / 104.2 & 216.4 / 104.0 & \textbf{229.6 } / 104.7 \\
        UniPose (bf16)~\cite{li2025unipose} & 553.0 / \textbf{93.6} & 551.2 / \textbf{92.1} & 551.0 / \textbf{92.8} & 549.8 / 96.2 & 551.3 / \textbf{93.7} \\
    \bottomrule
    \end{tabular}}
    \caption{\textbf{Quantitative results on the RPE benchmark.} Each entry reports MPJPE/PA-MPJPE scores (lower is better) across four text-description categories (\textit{Behavior}, \textit{Shape}, \textit{Outfit}, and \textit{Pose}). fp16 and bf16 represent half-precision and brain floating point 16 formats, respectively. Bold denotes the lowest (best) value in each column. \textit{Original} refers to results reported in the ChatPose~\cite{ChatPose} manuscript.}
    \label{tab:pose_estimation_comparison}
\end{table*}
\section{Open-Source release of refined RPE benchmark}
\label{section:open_source}
To resolve the technical reproducibility challenges, we publicly release carefully refined GT annotations. Specifically, we performed meticulous manual one-to-one verification, explicitly linking each benchmark example to the corresponding original 3DPW frame (\cref{fig:rpe_gt_cmp}, \textit{right}). Our refined annotations include essential information required for precise quantitative evaluations, such as SMPL parameters and 3D joint coordinates. By openly providing these refined ground truths, we eliminate the previously unavoidable manual matching step, enabling researchers to conveniently and consistently evaluate pose-aware MLLMs. The refined annotations are publicly available at \href{https://github.com/jjunsss/RPE-Refined}{\textbf{link}}.

\section{Experiment}
To demonstrate the practical utilization and reliability of our released RPE annotations, we conducted quantitative experiments using state-of-the-art pose-aware MLLMs, specifically ChatPose~\cite{ChatPose} and UniPose~\cite{li2025unipose}. 

\subsection{Setup details}
To evaluate original pose-aware MLLMs' capabilities for estimating human poses from reasoning-based textual queries, we employ a refined version of the RPE benchmark derived from the 3DPW dataset annotated by NeuralAnnot~\cite{Moon_2022_CVPRW_NeuralAnnot}. The refined annotations include two distinct types of GT joint representations: (1) \texttt{joint\_cam}, which contains 3D joint coordinates directly provided by the original 3DPW~\cite{von20183dpw} dataset; and (2) \texttt{fitted\_joint\_cam}, representing joint coordinates regressed from the corresponding SMPL mesh within a camera-centered 3D space. For consistency and accuracy, our experiments primarily utilize the \texttt{fitted\_joint\_cam} annotations, as these values inherently align with the SMPL~\cite{SMPL} mesh parameters. Nevertheless, both representations are included in our benchmark annotations to facilitate flexible usage according to specific user intentions or downstream task requirements.

\subsection{Evaluation details}
Although the original ChatPose~\cite{ChatPose} manuscript explicitly stated the use of the SMPL~\cite{SMPL}, the codebase utilizes the SMPL-X~\cite{SMPL-X}. To ensure a fair comparison, we follow the evaluation protocol described by \citet{sun2021monocular}, utilizing only 22 joints from the SMPL-X joint set, excluding the two hand joints. This 22-joint evaluation setup is also adopted by UniPose~\cite{li2025unipose}. We measure the accuracy of the estimated poses primarily using mean per joint position error (MPJPE). Additionally, procrustes-aligned MPJPE (PA-MPJPE) is computed to assess the similarity between predicted poses and GT poses, independent of global translation, rotation, and scaling effects. Furthermore, ChatPose~\cite{ChatPose} did not specify the inference precision, despite known performance variations associated with different floating-point types in LLM environments. Thus, we present results using both bf16 and fp16, while UniPose~\cite{li2025unipose} results utilize bf16.

\subsection{Quantitative results}
\cref{tab:pose_estimation_comparison} summarizes quantitative results on the RPE benchmark across four text-description categories: \textit{Behavior}, \textit{Shape}, \textit{Outfit}, and \textit{Pose}. Our results for ChatPose~\cite{ChatPose}, obtained using the refined GT annotations, closely align with the original values reported in~\cite{ChatPose} (see \textit{Original}), confirming the validity of our manual annotation refinement. Additionally, we provide previously unavailable quantitative results for UniPose~\cite{li2025unipose}. UniPose shows substantially higher MPJPE values compared to ChatPose across all categories, indicating that it struggles to accurately interpret natural-language queries requiring sophisticated semantic reasoning. While the UniPose authors fine-tuned their model to address this limitation, they have not publicly released this fine-tuned version. On the other hand, the lower PA-MPJPE results of UniPose demonstrate its strong intrinsic capability for pose representation. We hypothesize this discrepancy arises from fundamental differences in the way the two approaches utilize LLMs: ChatPose explicitly leverages general world-knowledge understanding, whereas UniPose primarily functions as a pose-specific generator. These distinctions likely affect their performance due to differences in dataset structure and training strategy.

\section{Conclusion}
\label{section:conclusion}
In this paper, we have systematically identified and addressed key limitations of the widely adopted reasoning-based pose estimation benchmark, specifically regarding reproducibility, reliability, and robustness. We resolved significant reproducibility issues by meticulously refining and publicly releasing accurate GT annotations aligned with the original 3DPW dataset, thereby facilitating consistent and rigorous quantitative evaluations. Furthermore, we thoroughly documented intrinsic benchmark-quality issues, including dataset redundancy, scenario imbalance, simplistic scenes, ambiguous queries, limited multi-person annotations, and information loss due to preprocessing. Our analysis provides a clear foundation for developing improved benchmarks that effectively address these limitations. In future work, we plan to propose an enhanced reasoning-based benchmark specifically designed to overcome these identified shortcomings. Moreover, by better integrating pose information as supportive cues for the global knowledge and reasoning capabilities inherent in LLMs, we anticipate advancing pose-aware multimodal models beyond current methods, such as ChatPose, towards more sophisticated, accurate, and semantically coherent human pose reasoning.

\section{Acknowledgment}
This research was supported by Brian Impact Foundation, a non-profit organization dedicated to the advancement of science and technology for all.

\clearpage %
\newpage
{
    \bibliographystyle{ieeenat_fullname}
    \bibliography{main}
}

\end{document}